\documentclass[10pt,twocolumn,letterpaper]{article}

\usepackage[usenames,dvipsnames]{xcolor}
\usepackage{paralist}
\usepackage{xspace}
\usepackage{comment}

\usepackage{cvpr}
\usepackage{times}
\usepackage{epsfig}
\usepackage{graphicx}
\usepackage{amsmath}
\usepackage{amssymb}
\usepackage{subfig}

\usepackage{algorithmic}
\usepackage{algorithm}
\usepackage{rotating}
\usepackage{diagbox}
\usepackage{balance}
\usepackage{booktabs}
\usepackage{paralist}

\newcommand{\myparagraph}[1]{\vspace{0.1em}\noindent\textbf{#1}}

\usepackage[pagebackref=true,breaklinks=true,letterpaper=true,colorlinks,bookmarks=false]{hyperref}

\newcommand{\figref}[1]{Fig.~\ref{#1}}

\cvprfinalcopy % *** Uncomment this line for the final submission

\def\cvprPaperID{16} % *** Enter the CVPR Paper ID here

\ifcvprfinal\fi
\begin{document}

%%%%%%%%% TITLE
\title{Unpaired Pose Guided Human Image Generation}

\author{Xu Chen \quad \quad Jie Song \quad \quad  Otmar Hilliges\\
AIT Lab, ETH Zurich\\
{\tt\small \{xuchen,jsong,otmarh\}@inf.ethz.ch}
}

\maketitle
%\thispagestyle{empty}

%%%%%%%%% ABSTRACT
\begin{abstract}
This paper studies the task of full generative modelling of realistic images of humans, guided only by coarse sketch of the pose, while providing control over the specific instance or type of outfit worn by the user. This is a difficult problem because input and output domain are very different and direct image-to-image translation becomes infeasible. We propose an end-to-end trainable network under the generative adversarial framework, that provides detailed control over the final appearance while not requiring paired training data and hence allows us to forgo the challenging problem of fitting 3D poses to 2D images.  
The model allows to generate novel samples conditioned on either an image taken from the target domain or a class label indicating the style of clothing (e.g., t-shirt). We thoroughly evaluate the architecture and the contributions of the individual components experimentally. Finally, we show in a large scale perceptual study that our approach can generate realistic looking images and that participants struggle in detecting fake images versus real samples, especially if faces are blurred.
\end{abstract}

%%%%%%%%% BODY TEXT
% !TEX root = ../egpaper_for_review.tex

\section{Introduction}
In this paper we explore full generative modelling of people in clothing given only a sketch as input. This is a compelling problem motivated by the following questions. First, humans can imagine people in particular poses, wearing specific types of clothing -- can machines learn to do the same? If so -- how well can generative models perform this task? This clearly is a difficult problem since the input, a sketch of the pose, and the output, a detailed image of a dressed person, are drastically different in complexity, rendering direct image-to-image translation infeasible. The availability of such a generative model, would make many new application scenarios possible: cheap and easy-to-control generation of imagery for e-commerce applications such as fashion shopping, 
%This finds application for example in creating imagery for e-commerce websites
or to synthesize training data for discriminative approaches in person detection, identification or pose estimation.

%\js{Here I have a bit concern that people will ask us to directly compare with those virtual try-on papers or Lassner(as the review). We might have to emphasize we could not compare with Lassner and for those virtual try-ons, they are very specific. We need to explain a bit why we need a single network to do both.}

%Generative modelling of the statistics of natural scenes for the synthesis of high-quality images is of central importance to computer vision. Deep learning methods have been used to translate images from one domain into another \cite{isola2017image,cyclegan}. More recently, attempts have been made to synthesize images \cite{chen2016synthesizing,Lassner:GeneratingPeople:2017,NIPS2017_6644} or videos \cite{chan2018everybody,wang2018video} of people in clothing, without having to manually specify material, lighting, dynamics of clothing and the interaction with the human body, which is feasible but cumbersome via the rendering pipeline.

\begin{figure}[!t]
 \includegraphics[width=0.5\textwidth]{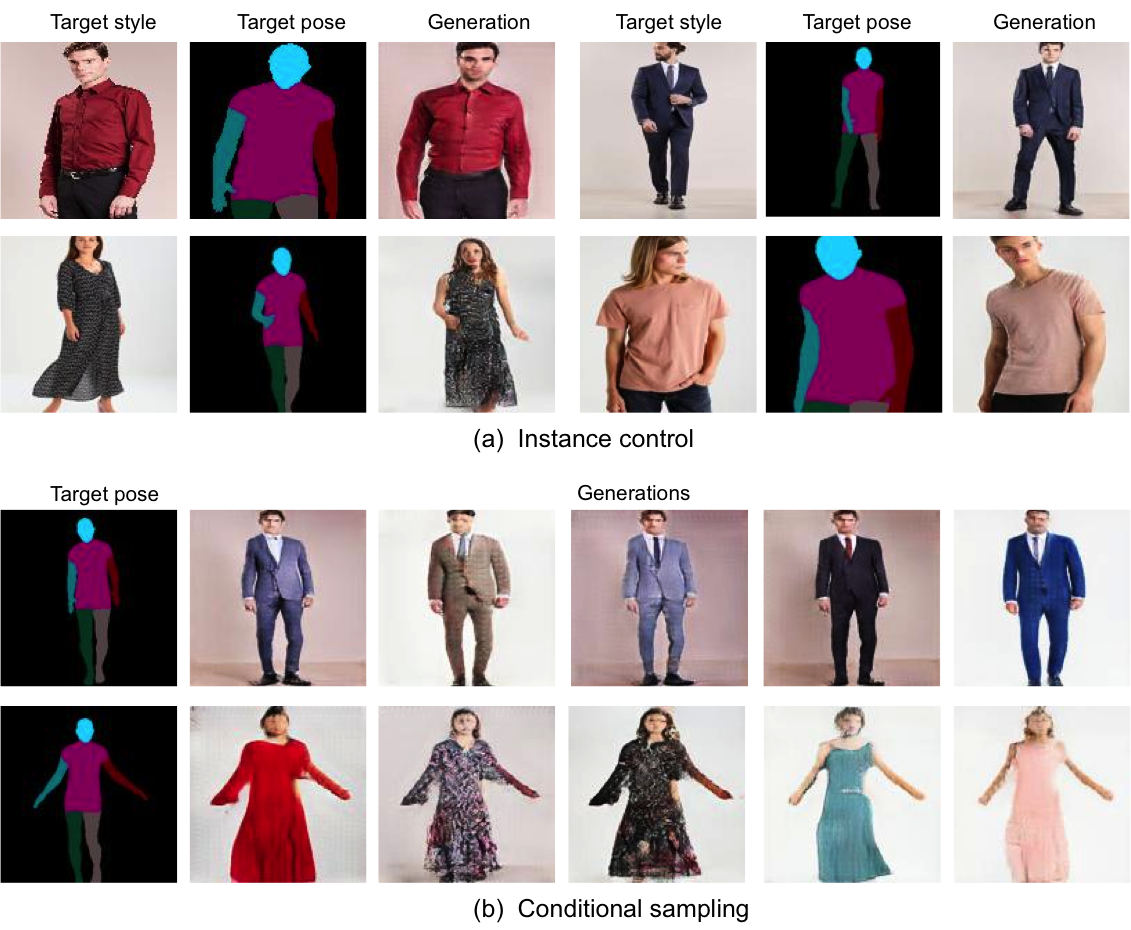}
    \caption{\textbf{Generating humans in clothing:} Our network takes a pose sketch as input and generates realistic images, giving users (a) instance control via an reference image, or (b) via conditional sampling, leading to images of variations of a class of outfits.}
    \label{fig:teaser}
\end{figure}

Existing approaches typically cast the problem of generating people in clothing as two-stage paired image-to-image translation task. Lassner \etal~\cite{Lassner:GeneratingPeople:2017} require pairs of corresponding input images, 3D body poses and so-called parsing images. While this allows for the generation of images with control over the depicted pose, such approaches do not provide any control over the type or even specific instance of clothing. However, many application scenarios would require control over all three. 

In this paper we propose a new task that, to the best of our knowledge, isn't covered by any related work:  we seek to  generate images of people in clothing with 
\begin{inparaenum}[(T1)]
\item control over the pose, and
\item exact instance control (the type of clothing), or 
\item conditional sampling via class label, and with sufficient variance (e.g., blue or black suit).
\end{inparaenum}
For example, provided an image of a person a model should be able to generate a new image of the person wearing a specific clothing item (e.g., red shirt) in a particular pose (Fig~\ref{fig:teaser} a). Or provided a pose and class label (e.g., dress) it should synthesize images of different variants of the clothing in target pose (Fig~\ref{fig:teaser} b). 
%This task definition is motivated by the following aspects: (i) academic interest: humans can imagine people in clothing - can machines learn to do the same? (ii) If so, how well can generative models perform this task? (iii) New applications: computationally cheap and easy-to-use pipelines could enable novel e-commerce applications (e.g., for virtual try-on). 

To tackle this problem, we furthermore contribute a new dataset, and a network architecture that greatly reduces the amount of annotations required for training which in turn drastically reduces human effort and thus allows for faster, cheaper and easier acquisition of large training data sets. Specifically, we only require \emph{unpaired} sets of 3D meshes with part annotations and images of people in clothing. These sets only need to contain roughly similar distributions of poses, allowing for reuse of existing datasets.

Furthermore, our work contributes on the technical level in that we propose a single end-to-end trainable network under the generative adversarial framework, that provides active control over  
\begin{inparaenum}[(i)]
\item the pose of the depicted person,
\item over the type of cloth worn via conditional synthesis, and even 
\item the specific instance of clothing item.
\end{inparaenum}
Note that in our formulation applying clothing is not performed via image-to-image transfer and hence allows for generation of novel images via sampling from a variational latent space, conditioned on a style label (e.g., t-shirt). Finally, our approach allows for the construction of a single inference model that makes the two-stage approach of prior work unnecessary. 

We evaluate our method qualitatively in an in-depth analysis and show via a large scale user study ($n=478$) that our approach produces images that are perceived as more realistic, indicated by a higher ``fool-rate'', than prior work. 
% !TEX root = ../egpaper_for_review.tex

\section{Related Work}
We consider the problem of generating images of people in clothing, with fine-grained user control. This is a relatively new area in the computer vision literature. Here we review the most closely related work and briefly summarize work that leverages deep generative models in adjacent areas of computer vision. 

\myparagraph{Generating people in clothing}
Synthesizing realistic images of people in different types of clothing is a challenging task and of great importance in many areas such as e-commerce, gaming and as potential source of training data for computer vision tasks. The computer graphics literature has dedicated a lot of attention to this problem including skinning and articulation of 3D meshes, simulation of physically accurate clothing deformation and the associated rendering problems~\cite{guan20102d,chen2016synthesizing,goldenthal2007efficient,pishchulin2012articulated,narain2012adaptive,kulkarni2015deep}. 
Despite much progress, generating photo-realistic renderings remains difficult and is computationally expensive. 

Circumventing the graphics pipeline entirely, image-based generative modeling of people in clothing has been proposed as emerging task in the computer vision and machine learning literature. 

One line of work \cite{han2017viton,Wang2018-oc, Raj2018-kv, Zanfir2018-fp} targets so-called virtual try-on, transferring garment appearance from a high-quality photograph, to the corresponding body part(s) in the destination 2D image. 

Another line of work, which is more related to ours, aims to generate human images with control over the exact body pose. \cite{Lassner:GeneratingPeople:2017, NIPS2017_6644,ma2018disentangled, Si2018-xk, Chan2018-rr, Bem2019-ny,Esser2018-da,Pumarola2018-ty} generate images at the pixel level with variants of encoder-decoder architectures. \cite{Neverova2018-dh, Siarohin2018-wq, Balakrishnan2018-yl,Dong2018-rt} further incorporate spatial warping to form human images in novel poses.
However, the above methods rely on detailed annotations of natural human images, e.g. body poses \cite{NIPS2017_6644,ma2018disentangled,Si2018-xk,Chan2018-rr, Bem2019-ny, Esser2018-da, Pumarola2018-ty,Neverova2018-dh, Dong2018-rt,Siarohin2018-wq,Balakrishnan2018-yl} or clothing segmentations \cite{Lassner:GeneratingPeople:2017, Dong2018-rt}, which are non-trivial to obtain. In contrast, our work aims to synthesize human images without these annotations. The only required training data for our network are \emph{unpaired} sets of images of people in clothing and 3D meshes with part annotations. Being trained unsupervised, our method still provides control over both pose and appearance of the generated images. 

% The most related work in spirit to ours is~\cite{Lassner:GeneratingPeople:2017}, proposing a generative model of clothed people, learned directly from images. Their model allows to randomly generate images of people from a learned latent space in again a two-stage process. Our method differs significantly from~\cite{Lassner:GeneratingPeople:2017} in the following aspect: 
% \begin{inparaenum}[(i)]
% \item we do not need any paired data for training, which are non-trivial to obtain;
% \item we do not need 3D joint annotations for the real images;
% \item we do not need any human parsing annotations;
% \item we can actively control the type of clothing (e.g., t-shirt vs dress-shirt) or even the instance of the clothing (e.g., a blue t-shirt) while providing the ability to generate new instances by sampling from the latent space.
% \end{inparaenum}
% Finally, our model consists of a single end-to-end trainable architecture and can be trained via data taken from different domains as long as the poses depicted in each roughly correspond (i.e., if 3D meshes show torsos, only the target images need to show similar 2D content). 

%%%%%%%%%%%%
%% DOUBLECOL FIGURES NEED TO BE INCLUDED TWO PAGES AHEAD
%%%%%%%%%%%%
\begin{figure*}[t!]
\begin{center}
 \includegraphics[width=0.95\linewidth]{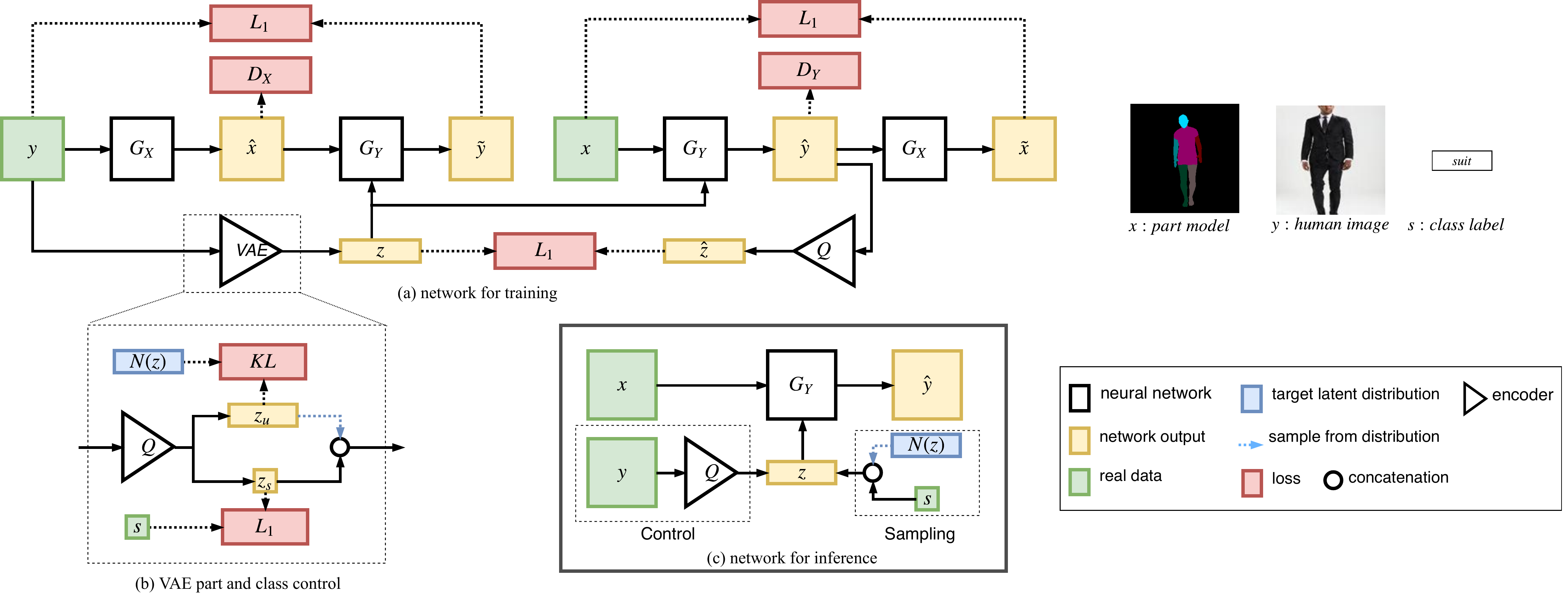}
 \end{center}
\caption{\textbf{Schematic architecture overview.} (a): network pipeline for training; (b): inset shows encoder in detail, consisting of a VAE and class label code; (c) the inference network generates synthetic images $\hat{y}$, given part model $x$ as input and using either a style class label $s$ or style depiction $y$ to condition the generation process. }
\label{fig:network}
\end{figure*}
%%%%%%%%%%%%
%% DOUBLECOL FIGURES NEED TO BE INCLUDED TWO PAGES AHEAD
%%%%%%%%%%%%

\myparagraph{Deep generative models}
Deep generative models that can synthesize novel samples have lately seen a lot of attention in the machine learning and computer vision literature. Especially generative adversarial networks (GANs)~\cite{goodfellow2014generative,salimans2016improved,DBLP:journals/corr/RadfordMC15,mao2017least} and variational autoencoders (VAEs)~\cite{kingma2013auto} have been successfully applied to the task of image synthesis. 
A particular problem that needs to be addressed in real tasks is that of control over the synthesis process. In their most basic form neither GANs nor VAEs allow for direct control over the output or individual parameters. 
To address this challenge a number of recent studies have investigated the problem of generating images conditioned on a given image or vector embeddings~\cite{kingma2014semi,isola2017image,yan2016attribute2image} such that they predict $p(\vec{y}|\vec{x})$, where $\vec{x}$ is a reference image or similar mean of defining the desired output (e.g., one-hot encoded class labels).

\myparagraph{Image-to-image translation}
Generating people in clothing given a target pose can be regarded as an instance of image-to-image translation problem. In~\cite{isola2017image}, automatic image-to-image translation is first tackled based on conditional GANs. While conditional GANs have shown promise when conditioned on one image to generate another image, the generation quality has been attained at the expense of lack of diversity in the translated outputs. To alleviate this problem, many works propose to simultaneously generate multiple outputs given the same input and encourage them to be distinct~\cite{bicyclegan,chen2017photographic,bansal2017pixelnn}.

In~\cite{bicyclegan}, the  ambiguity  of  the  many-to-one mapping  is  distilled  into  a  low-dimensional  latent  vector, which can be randomly sampled at test time and encourages generation of more diverse results.
To train all of the above methods, paired training data is needed. 
However, for many tasks, paired information will not be available or will be very expensive to obtain.  
In~\cite{cyclegan}, an approach called CycleGAN to translate an image from a source domain X to a target domain Y in the absence of paired examples has been introduced. Similar approaches have been proposed to enforce domain translation with cycle consistency loss~\cite{kim2017learning,yi2017dualgan}.

In our paper, we build on this prior work but leverage a formulation that gives control over individual dimensions when generating images and the capability to generate entirely novel samples via incorporation of a variational auto-encoder that can be conditioned on a particular pose. A couple of concurrent works also recognize these limitations and propose architectures for the somewhat easier task of unsupervised multi-modal image domain translation~\cite{almahairi2018augmented,huang2018multimodal,lee2018diverse}. We show that our architecture yields comparable results even though this task is not the main focus of our work.
% !TEX root = ../egpaper_for_review.tex

\section{Method}\label{sec:method}

% \begin{figure*}
% \begin{center}
%  \includegraphics[width=0.9\linewidth]{pics/pipeline.pdf}
%  \end{center}
%     \caption{\textbf{Network architecture.} 2D source views are encoded and the latent code is explicitly rotated before a decoder network predicts a point cloud in the target view. Dense correspondences are attained via perspective projection. Bilinear sampling preserves local details in the final view. All operations are differentiable and trained end-to-end without ground-truth depth or flow maps.}
%     \label{fig:pipeline}
% \end{figure*}

% Naming scheme:
% \begin{itemize}
% \item $\vec{x}$: part model (2D projection of labeled 3D mesh)
% \item $\vec{y}$: human image 
% \item $\vec{z}$: latent code
% \item $s$: class label (clothing)
% \item $\hat{\vec{x}}$: synthetic part model (2D)
% \item $\hat{\vec{y}}$: synthetic human image
% \item $\tilde{\vec{x}}$: reconstructed part model $G_X(\hat{\vec{y}})$
% \item $\tilde{\vec{y}}$: reconstructed human image $G_Y(\hat{\vec{x}})$

% \end{itemize}

\newcommand{\inputhumanimage}{source human image}
\newcommand{\synthetichumanimage}{generated human image}
\newcommand{\syntheticpartmodel}{generated part model}
\newcommand{\inputpartmodel}{source part model}

To conditionally synthesize images of people in clothing, while providing user control over the generation process,  we propose a network combining the benefits of generative adversarial networks with cycle consistency and variational autoencoders in a single end-to-end trainable network. The aim is to generate high-quality images and to allow for the conditional generation of samples where the appearance is controlled via either an image (e.g., a specific suit) or a class label indicating the style of clothing (e.g., suits in general). Importantly our architecture can be trained via unpaired training data. That is 3D poses and real images do not need to correspond directly. 
The architecture, illustrated in~\figref{fig:network} and dubbed Unpaired Pose-Guided GAN (UPG-GAN) combines elements of the GAN and VAE framework with several novel consistency losses to enable the fine-grained appearance control. 
In this section, we briefly introduce the basics of CycleGANs and VAEs and then detail our proposed architecture and training scheme.

\subsection{Preliminaries: CycleGAN and VAE}
To enable the desired unpaired training scheme, we build our framework on the basis of CycleGAN~\cite{cyclegan}. In the CycleGAN framework two generators $G_Y:X\rightarrow Y$ and $G_X:Y\rightarrow X$ translate images from one domain to another and attempt to fool the corresponding discriminators $D_Y$ and $D_X$, classifying samples into real and synthetic data. This is expressed via the following minimax game:
% \begin{flalign}
% \min_{G_X,G_Y}\max_{D_X,D_Y} &( \mathcal{L}_{GAN}(G_X,D_X,Y,X) \\ &+ {L}_{GAN}(G_Y,D_Y,X,Y)),
% \end{flalign}

{\footnotesize
  \setlength{\abovedisplayskip}{3pt}
  \setlength{\belowdisplayskip}{\abovedisplayskip}
  \setlength{\abovedisplayshortskip}{0pt}
  \setlength{\belowdisplayshortskip}{3pt}
	\begin{align}
	\min_{G_X,G_Y}\max_{D_X,D_Y}  ( \mathcal{L}_{GAN}(G_X,D_X,Y,X) + \mathcal{L}_{GAN}(G_Y,D_Y,X,Y)),
	\end{align}
}%
where $\mathcal{L}_{GAN}$ is the standard GAN loss:

{\footnotesize
  \setlength{\abovedisplayskip}{3pt}
  \setlength{\belowdisplayskip}{\abovedisplayskip}
  \setlength{\abovedisplayshortskip}{0pt}
  \setlength{\belowdisplayshortskip}{3pt}
  % \begin{flalign}
% \mathcal{L}_{GAN}(G,D,X,Y) = 
% \mathbb{E}_{x\sim p(x)}[\log{D(x)}] + \mathbb{E}_{y\sim p(y)}[\log{(1 -D(G(y)))}].
% \end{flalign}
	\begin{flalign}
	\mathcal{L}_{GAN} = 
	\mathbb{E}_{x\sim p(x)}[\log{D(x)}] + \mathbb{E}_{y\sim p(y)}[\log{(1 -D(G(y)))}].
	\end{flalign}
}%
To prevent $G_X$ and $G_Y$ from neglecting the underlying information of the input source images, a cyclic reconstruction loss $\mathcal{L}_{cyc}$ is added: 

{\footnotesize
  \setlength{\abovedisplayskip}{3pt}
  \setlength{\belowdisplayskip}{\abovedisplayskip}
  \setlength{\abovedisplayshortskip}{0pt}
  \setlength{\belowdisplayshortskip}{3pt}
	\begin{align}
	\mathcal{L}_{cyc}(G_X,G_Y,X,Y) &= \mathbb{E}_{x\sim p(x)}[\left \| G_X(G_Y(x)) - x \right \|_1] \notag &\\ 
	&+ \mathbb{E}_{y\sim p(y)}[\left \| G_Y(G_X(y)) - y \right \|_1 ].
	\end{align}
}%
% The intuition is that by demanding the recovery of the input image from the synthetic image, $G$ is encouraged to preserve the structure underlying in the input image. 
The overall objective of CycleGAN is then given by:

{\footnotesize
  \setlength{\abovedisplayskip}{3pt}
  \setlength{\belowdisplayskip}{\abovedisplayskip}
  \setlength{\abovedisplayshortskip}{0pt}
  \setlength{\belowdisplayshortskip}{3pt}
	\begin{align}
	\mathcal{L}_{CycleGAN}(G_X, G_Y, D_X, D_Y) &= \mathcal{L}_{GAN}(G_X,D_X, Y, X)  \notag\\
	&+\mathcal{L}_{GAN}(G_Y,D_Y, X, Y) \notag\\
	&+ \mathcal{L}_{cyc}(G_X,G_Y,X,Y).
	\end{align}
}%
This is optimized by alternating between maximizing the generator and minimizing the discriminator objectives.

Furthermore, we would like to endow the model with the ability to stochastically generate samples of natural human images $\hat{y}$ (to attain natural variation in appearance), given a specified pose. For this task, we leverage VAEs so that a latent code $z$ can be sampled from a prior distribution $\mathcal{N}(z)$. In the VAE setting, $\mathcal{N}(z)$ is commonly chosen to be an isotropic normal distribution $\mathcal{N}(0, 1)$ with zero mean and unit variance. Applying the re-parametrization trick the corresponding KL-Divergence ($D_{KL}[\mathcal{N}(\mu(y), \sigma(y)^2) \Vert \mathcal{N}(0, 1)] $) is minimized during training to regularize the latent distribution to be close to $\mathcal{N}(z)$, and $\mu$ and $\sigma$ are encoded via the image $y$:
% \footnotesize
% \begin{flalign}
% D_{KL}[\mathcal{N}(\mu(y), \sigma(y)^2) \Vert \mathcal{N}(0, 1)] = \frac{1}{2} \sum_k ( \exp(\sigma(y)^2) + \mu^2(y) - 1 - \sigma(y)^2 )  ,
% \end{flalign}
% \normalsize

{\footnotesize
  \setlength{\abovedisplayskip}{3pt}
  \setlength{\belowdisplayskip}{\abovedisplayskip}
  \setlength{\abovedisplayshortskip}{0pt}
  \setlength{\belowdisplayshortskip}{3pt}
	\begin{flalign}
	D_{KL} = \frac{1}{2} \sum_k ( \exp(\sigma(y)^2) + \mu^2(y) - 1 - \sigma(y)^2 )  .
	\end{flalign}
}%

\subsection{Unpaired Pose Guided GANs (UPG-GAN)}\label{sec:upg-GAN}
We now discuss (1) how to endow the base CycleGAN architecture with control over the specific instance of clothing (provided via a reference image) and (2) how to allow for conditional generation of a diverse set of images via sampling where the style is controlled via a class label $s$.\\ 

\myparagraph{Instance control.}
To allow for control over the clothing in the output image, we first introduce an additional encoder $Q:Y\rightarrow Z$ (dashed box in \figref{fig:network} (a)) to extract information from the input human image $y$ into a latent code $z$. This code then serves as a guidance to reconstruct a human image $\tilde{y}=G_Y(\hat{x},z)$ from the generated synthetic part model $\hat{x}=G_X(y)$. If $z$ is simply fed to $G_Y$ the generation process of $\hat{y}$ suffers from mode-collapse due to information hiding. During training $G_X$ embeds a nearly imperceptible, high frequency signal into $\hat{x}$. At inference time $x$, the input to $G_Y$, is void of the hidden signal and generation of $\hat{y}$ converges to a single mode. This problem is especially severe when one domain, in our case the human images $Y$, represents significantly richer and more diverse information than the other, in our case the part models $X$ (see \figref{fig:network}). To prevent this, $\hat{y}$ is enforced to reflect the information encoded in $z$ via introduction of a latent code consistency loss: 

{\footnotesize
\begin{flalign}
\label{eqt:reg_loss}
\mathcal{L}_{c}(G_Y, Q)= \mathbb{E}_{x\sim p(x), y\sim p(y)} [\left \| Q( G_Y(x, z ) ) - z \right \|_1].
\end{flalign}
}%
Here the encoder $Q$ produces a style code $\hat{z}=Q(\hat{y})$ from $\hat{y}$.\\ 

\myparagraph{Conditional sampling.}
In cases for which only the type of clothing matters but diverse image generation is desired (i.e., to triage different designs) we extend our architecture to accept a class label $s$ for implicit guidance.   
To be able to generate varied images via sampling from a prior distribution, we propose a scheme similar to the VAE framework, albeit without an explicit decoder (see \figref{fig:network}, inset). 
Conditioning with a specific type of clothing requires disentanglement of style-type information from other dimensions of the latent code such as color and texture of the clothing. 
Following the approach in ss-InfoGAN \cite{spurr2017ecml}, we leverage easy to attain style class labels $s$ while we do not provide labels for any of the other dimensions. The latent code $z$ is then decomposed into a supervised part $z_s$, controlling the style class, and an unsupervised part $z_{u}\sim Q(z|y)$, where $z_s\cup z_{u}=z$. The following regularization term encourages the VAE's encoder $E$ to learn a disentangled representation
\begin{flalign}
\mathcal{L}_s(Q_{s},Y, S) = \mathbb{E}_{y\sim p(y)} [\left \| Q_s(y) - s \right \|_1] ,
\end{flalign}
via enforcing similarity between $z_s$ and its corresponding ground truth label $s$. The inference network (\figref{fig:network}, c) can then be fed with labels $s$ to control style classes while producing samples with sufficient variability (\figref{fig:teaser}, b). %See \figref{fig:label} for an illustration. 

% To control the type of generated images and to guide the training process of an unified model for multiple classes, the class label $s$ of each image is used as semi-supervision by concatenating it to the code generated by the encoder/ variational encoder, which is similar to the method in \cite{cond-cyclegan}. However the network tends to ignore this information. In order to prevent this and also enable the network to infer the class during testing phase,  an encoder $E_c$ is trained to predict the class of real images and is used to infer the class of generated images as a guidance to the generator.  $E_c$ is trained simultaneously with the whole network by optimizing the classification loss given by 
% \begin{flalign}
% \label{eqt:class_loss}
% L_{class}(E_c)= \mathbb{E}_{ y\sim p_{data}(y)} [\left \| E_c( y) - c \right \|]
% \end{flalign}
% , where $c$ is the ground truth class label. \todo{Explain the reason why using L1 loss instead of cross entropy loss}.  The mutual information loss is extended by a new term
% \begin{flalign}
% \label{eqt:class_info_loss}
% L_{info, class}(G_A, E_c)= \mathbb{E}_{x\sim p_{data}(x), y\sim p_{data}(y)} [\left \| E_c( G_A(x, E_c(y) ) ) - E_c(y) \right \|]
% \end{flalign}

% \begin{figure}
% \includegraphics[width=\textwidth]{pics/class_encoder.png}
% \caption{Class Encoder}
% \label{fig:vari_enc}
% \end{figure}

\subsection{Training Schedule}
\noindent Our overall objective function: 
{\footnotesize
\begin{flalign}
\mathcal{L}_{UPG-GAN}
= \mathcal{L}_{GAN}
+ \mathcal{L}_{cyc} 
+ \mathcal{D}_{KL} 
+ \mathcal{L}_{c}
+ \mathcal{L}_{s} ,
\label{eqt:overall_loss}
\end{flalign}
}%
is optimized by the iterative training procedure given as pseudo-code in Alg.\ref{alg:training}. At the beginning of each iteration, one part model $x$ and one human image $y$ are randomly drawn from the corresponding datasets, and the ground truth label $s$ for $y$ is read, if available. The VAE encodes $y$ to get the latent code $z$. Synthetic samples $\hat{x}$ and $\hat{y}$ are generated from $x,y$ and $z$, and at the same time a latent code $\hat{z}$ is extracted from $\hat{y}$. The last step of the forward pass is to reconstruct $\tilde{y}$ from $\hat{x}$ and $\tilde{x}$ from $\hat{y}$ respectively. The overall loss is calculated and back-propagated through the network to update the weights.

\newcommand{\fake}{\hat}
\newcommand{\rec}{\tilde}
\newcommand{\real}{}
\newcommand{\pred}{}
\newcommand{\predfake}{\fake}
\newcommand{\groundtruth}{}

\begin{algorithm}
\caption{Unpaired Pose Guided GAN}
\label{alg:training}
\begin{algorithmic}
	\STATE $x$ part model, ${X}$ part model dataset
    \STATE $y$ human image, ${Y}$ human image dataset
	\STATE $s$: class label (clothing), ${z}$: latent code
% \STATE $\hat{x}$: synthetic part model$\hat{{y}}$: synthetic human image
% \STATE $\tilde{{x}}$: reconstructed part model $G_X(\hat{{y}})$
% \STATE $\tilde{{y}}$: reconstructed human image $G_Y(\hat{{x}})$

    \FOR{$number\  of\  training\  epochs$}
          % forward pass
          \STATE $\real{x} \leftarrow {X}, \real{y},\groundtruth{s} \leftarrow {Y}$
          \STATE $z \leftarrow Q(\real{y})$ where $z = z_s \cup z_{u}$
%           \STATE $\pred{\mu}, \pred{\sigma}, \pred{s} \leftarrow E(\real{y})$
%           \STATE $\pred{z} \sim \mathcal{N}(\pred{\mu}, \pred{\sigma}^2)$

          \STATE $\fake{x} \leftarrow G_X(\real{y})$, $\fake{y} \leftarrow G_Y(\real{x},\pred{z})$
          \STATE $\rec{x} \leftarrow G_X(\fake{y})$, $\rec{y} \leftarrow G_Y(\fake{x},\pred{z})$, $\predfake{z}\leftarrow Q(\fake{y})$ %$\predfake{\mu},\predfake{s} \leftarrow E(\fake{y})$
%  		  \STATE $\predfake{\mu},\predfake{c} \leftarrow E(\fake{y})$
%           \STATE $\pred{d_x} \leftarrow D_X(\real{x})$, $\pred{d_y} \leftarrow D_Y(\real{y})$ \todo{mention image pool?}
%           \STATE $\predfake{d_x} \leftarrow D_X(\fake{x})$, $\predfake{d_y} \leftarrow D_Y(\fake{y})$ 
          
          % calculate loss
          \STATE $ \mathcal{L}_{GAN}^D \leftarrow (D(\real{x}) - 1)^2 + (D(\real{y}) - 1)^2 + D(\fake{x})^2 + D(\fake{y})^2 $
          \STATE $ \mathcal{L}_{GAN}^G \leftarrow (D(\fake{x}) - 1)^2 + (D(\fake{y}) - 1)^2$ 
          \STATE $ \mathcal{L}_{rec} \leftarrow  \left \| \rec{x} - \real{x} \right \|_1 + \left \| \rec{y} - \real{y} \right \|_1$
%           \STATE $\mathcal{D}_{KL} \leftarrow -0.5 * (1 +\log(\pred{\sigma}^2) - \pred{\mu}^2 - \pred{\sigma}^2)$
          \STATE $ \mathcal{L}_{c} \leftarrow  \left \| \predfake{z} -  \pred{z} \right \|_1$, $ \mathcal{L}_{s} \leftarrow  \left \| \groundtruth{s} - \pred{z_s} \right \|_1$
%           \STATE $ \mathcal{L}_{c} \leftarrow  \left \| \real{c} - \pred{c} \right \|_1$ \todo{notation for ground truth and prediction}
		  % back propagation
		  \STATE $\theta_{D_X},\theta_{D_Y} \leftarrow Adam(\mathcal{L}_{GAN}^D)$
          \STATE $\theta_{Q} \leftarrow Adam(\mathcal{L}_{GAN}^G + \mathcal{L}_{rec} + \mathcal{L}_{s} + \mathcal{D}_{KL})$
		  \STATE $\theta_{G_X} \leftarrow Adam(\mathcal{L}_{GAN}^G + \mathcal{L}_{rec})$ 
		  \STATE $\theta_{G_Y} \leftarrow Adam(\mathcal{L}_{GAN}^G + \mathcal{L}_{rec} + \mathcal{L}_{c})$
% 		  \STATE $\theta_{G_Y} \leftarrow Adam(\mathcal{L}_{GAN}^G + \mathcal{L}_{rec} + \mathcal{L}_{LR})$

    \ENDFOR
\end{algorithmic}
\end{algorithm}

% \begin{enumerate}
% \item generate z from realB.
% \item generate fakeA from realB and fakeB from realA and z (and c).
% \item generate recB from fakeA + z (and c) and recA from fakeB.
% \item generate zpred (and cpred )from fakeB.
% \item (optimize E with classification loss).
% \item optimize DA and DB with fakeA, fakeB and realA, realB.
% \item optimize GA, GB and E with reconstruction loss and gan loss.
% \item optimize GA with mutual information loss.
% \end{enumerate}
% !TEX root = ../egpaper_for_review.tex

\section{Experiments}
Evaluating generative models is a difficult task since the main goal, that of synthesizing novel samples, implies that no ground-truth information is available. Prior work on generating images of people in clothing has reported reconstruction accuracy. However, in our work this is not possible since we do not train on pairs of images. Furthermore, for the final task the two most important aspects are degree of control over the content and the final image quality. For these reasons we report mostly qualitative results but compare the various aspects of our proposed architecture in an ablative manner against the underlying building blocks (e.g., CycleGAN only). Finally, we report results from a large scale user study in which we asked participants to discriminate between randomly sampled real and fake images.

\subsection{Dataset}
One of the contributions in our work is the removal of the requirement for a task specific dataset. Training of the proposed architecture only relies on two separate sources of data: images of body part models
%with known 3D pose (e.g., from the 3D pose estimation literature) 
and images of real people wearing varying types of clothing. 
To attain samples of \emph{part models} we use the dataset of~\cite{Lassner:GeneratingPeople:2017}. To attain the real \emph{human images}, we crawled 1500 images from an online fashion store\footnote{\url{https://www.zalando.ch/}}, including t-shirts, dress-shirts, dresses and suits. The label $s$ was extracted from the online shop's categorization. To ensure rough correspondences between the body models and the images, we separated the datasets into those depicting full bodies and upper bodies only. Importantly these two data sources need not be directly paired and hence there is no need to fit the 3D body models to the 2D imagery. The dataset and code for network training and inference are released \footnote{\url{https://github.com/cx921003/UPG-GAN}}. %\oh{say something about $s$}.

% among which 580 are images of models wearing t-shirts, 290 are wearing dress-shirts, 240 are wearing dresses and 370 are wearing suits. 
\begin{figure}
	\includegraphics[width=\linewidth]{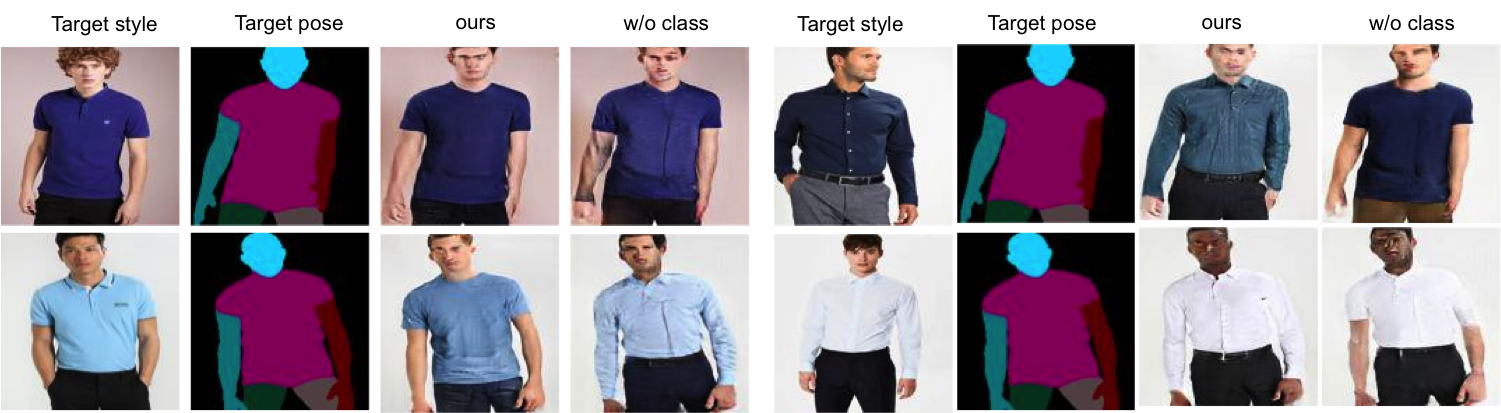}
    \caption{Several images produced by ours and w/o $\mathcal{L}_s$, conditioned on target human images and poses. Each four images in a row form a group. From left to right in a group: target style image, target pose, our result and  w/o $\mathcal{L}_s$.}
    \label{fig:label_control}
\end{figure}

\subsection{Implementation Details}
We set the network input to a fixed size of $128\times128$ for computational reasons. The two generators $G_X,G_Y$ (cf. Fig. \ref{fig:network}) share the same architecture and we use a standard Downsampling-ResNetChain-Upsampling configuration.  The VAE block and the classifier share the convolutional layers of a ResNet~\cite{he2016deep} architecture. 
During training, the learning rate is set to $0.00006$.  The weights of loss terms are set to $\omega_{s} = 1$, $\omega_{c}=10$, $\omega_{KL} = 0.01$, $\omega_{cyc} = 10$ and $\omega_{GAN} = 1$. 
\subsection{Ablation Study}
To understand the effect of each component that we add to the base CycleGAN architecture, we conduct an ablation study. We contribute four novel aspects, namely the clothing encoder, the latent code consistency loss, the KL divergence loss and the class supervision loss. The clothing encoder and KL divergence loss are necessary to perform instance control (via source image) and conditional sampling (via class label) respectively. Therefore we focus our study on the latent code consistency loss and the class supervision loss. We train without the latent code consistency and its corresponding loss $\mathcal{L}_{c}$ and a network without $\mathcal{L}_{s}$ and evaluate their performance on both instance control and conditional sampling, compared to our full network. Note that we cannot compare directly to the base architecture (without both) since it has severe problem of mode collapse and also it fails to provide control over the depicted outfit.
% We also provide the performance of CyleGAN on instance control as a reference. Note that CycleGAN can not be expected to be able to perform instance control (given a specific person image) and also conditional sampling hence the corresponding result is omitted.

% We now report experiments on our architecture when integrating the CycleGAN with the VAE component. Furthermore, to assess the effect of adding class labels and a corresponding loss term we train two variants of our architecture on the union of the t-shirt and dress-shirt datasets. The first variant is trained with supervised regression loss (ours) and the second without (CycleGAN+VAE).

\begin{figure}
	\includegraphics[width=\linewidth]{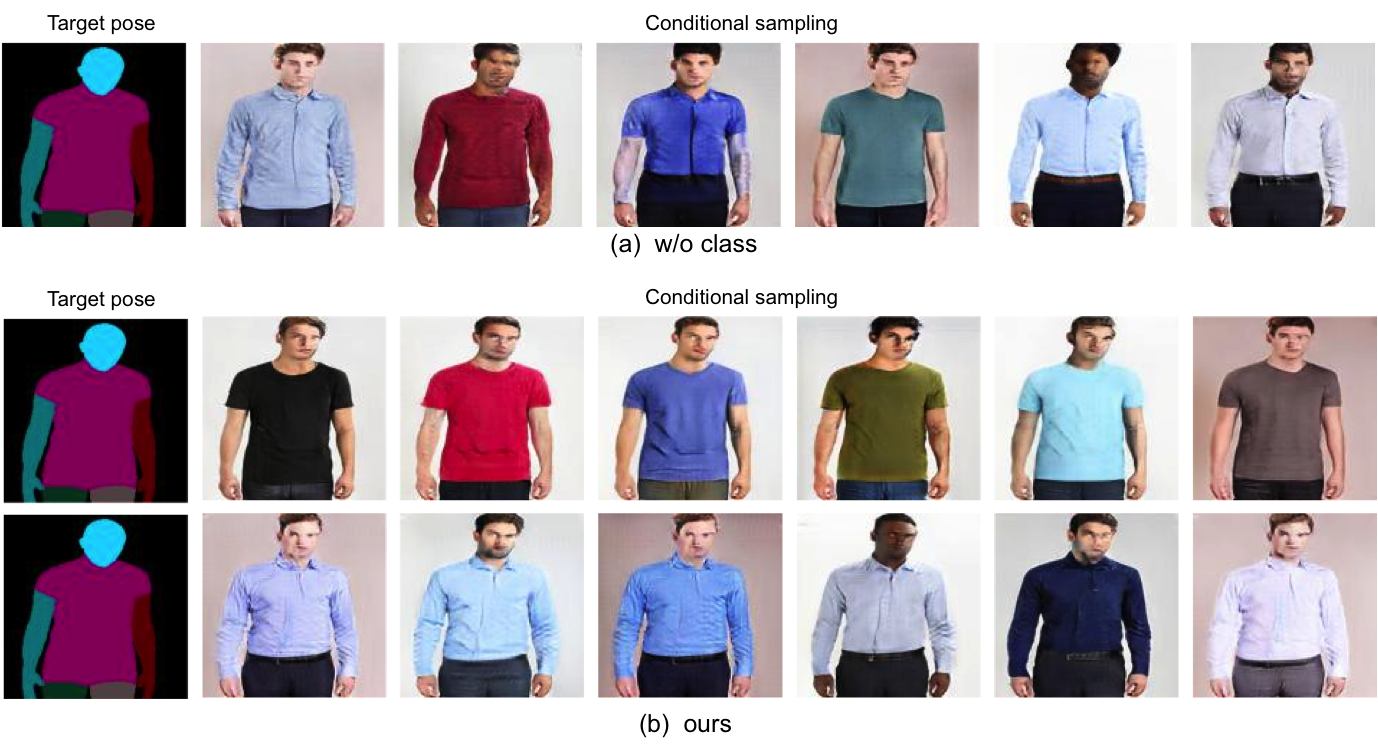}
    \caption{Conditional sampling comparison. (a): without $\mathcal{L}_s$ loss. (b): ours with $\mathcal{L}_s$.}
    \label{fig:sampling_compare}
\end{figure}

% \begin{figure}
% \centering
%      \subfloat[Sampling without using labels]{\includegraphics[width=\linewidth]{pics/no_label_sample.png}\label{fig:no_label_sample}} \\
%      \subfloat[Sampling with labels]{\includegraphics[width=\linewidth]{pics/label_sample.png}\label{fig:label_sample}}
% \caption{Synthetic samples generated a) with and b) without style label.}
% \label{fig:label_sampling}
% \end{figure}

\myparagraph{Setup:} During evaluation of instance control the realism of the generated human images and the similarity of style between the generation and the target are considered. To measure the realism we use a pre-trained faster-RCNN \cite{renNIPS15fasterrcnn} to detect people in the generated images and report the detection accuracy and average confidence. To evaluate if the target style persists in the output, we use a pre-trained person re-identification network \cite{xiaoli2017joint}. If the generated human image has a similar style to the target, the re-identification network should be able to re-identify it in the output image.
To evaluate conditional sampling, we sample 20 human images for each input body part model and again compute the person detection accuracy and the average confidence. Moreover, we randomly pick 19 sample pairs for each input body part model, and measure the diversity of the generation with the average LPIPS score as proposed in \cite{bicyclegan}. 

\myparagraph{Results:} We first discuss the \textbf{instance control} results as shown in Tab.~\ref{tab:ablation_study_control}. As indicated by the re-identification confidence, we can see that without the latent code consistency  (w/o $\mathcal{L}_c$) the network cannot obey the target style image well. The classification loss $\mathcal{L}_s$ helps to slightly improve the style-preservation. In addition we can see that both the latent code consistency and the classification loss help to improve the realism, reflected by the decrease in detection accuracy and confidence when either of these two is absent. We can see from Fig.~\ref{fig:label_control} that even though the version without $\mathcal{L}_s$ loss can produce satisfactory human images and also can keep the color correctly, it does not maintain the clothing type.

We now analyze the results in terms of \textbf{conditional sampling}. The low diversity score (0.0001) for the network trained without the latent code consistency loss $\mathcal{L}_c$ indicates that the provided latent codes are ignored during inference. By adding the latent code consistency loss, our full network can produce diverse samples with a much higher diversity score of 0.0700. Notably, the network without $\mathcal{L}_s$ achieves an even higher diversity score. However, this is due the unrealistic samples produced by this network, which is confirmed in the qualitative results and is also reflected in the low person detection scores. Fig.~\ref{fig:sampling_compare} shows that without $\mathcal{L}_s$ the network can produce diverse results but these are not consistent with the desired the type of clothing.

\begin{table}[]
\centering
\begin{tabular}{@{}cccc@{}}
\toprule
          & \begin{tabular}[c]{@{}c@{}}Accuracy\\ (control)\end{tabular} & \begin{tabular}[c]{@{}c@{}}Confidence\\ (control)\end{tabular} & \begin{tabular}[c]{@{}c@{}}Re-ID\\ (control)\end{tabular} \\  \midrule
w/o $\mathcal{L}_c$    & 61.4\% & 86.6\% & 0.382 \\
w/o $\mathcal{L}_s$ & 90.7\% & 91.9\% & 0.642  \\
Ours      & \textbf{94.3\%} & \textbf{94.4\%}  & \textbf{0.665}  \\ \bottomrule

\end{tabular}
\caption{Ablation study for instance control.}
\label{tab:ablation_study_control}
\end{table}

\begin{table}[]
\centering
\begin{tabular}{@{}ccccccc@{}}
\toprule
          &  \begin{tabular}[c]{@{}c@{}}Accuracy\\ (sample)\end{tabular} & \begin{tabular}[c]{@{}c@{}}Confidence\\ (sample)\end{tabular} & \begin{tabular}[c]{@{}c@{}}Diversity\\ (sample)\end{tabular} \\ \midrule
w/o $\mathcal{L}_c$    & 61.4\% & 86.4\%  & 0.0001 \\
w/o $\mathcal{L}_s$   & 80.7\%& 81.3\%  & \textbf{0.1240}  \\
Ours        & \textbf{93.9\%}  & \textbf{94.2\%}  & 0.0700 \\ \bottomrule
\end{tabular}
\caption{Ablation study for conditional sampling.}
\label{tab:ablation_study_sampling}
\end{table}

\subsection{Latent Space Visualization}
\begin{figure}[]
	\includegraphics[width=0.5\textwidth]{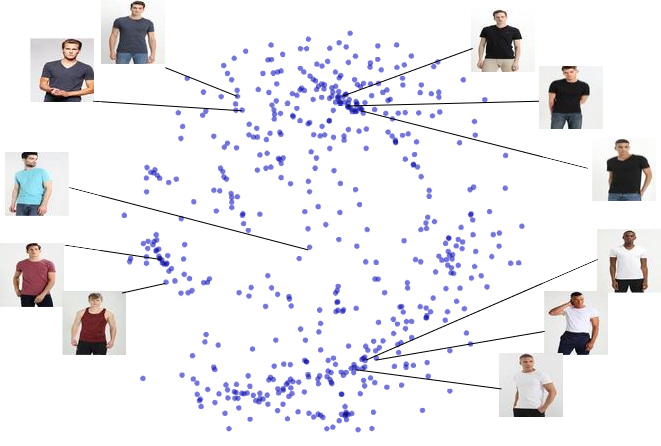}
    \caption{Visualization of the latent space.}
    \label{fig:tsne}
\end{figure}

\figref{fig:tsne} visualizes the learned latent space, indicating that the latent codes indeed capture clothing information. We encode all of our training images that depict t-shirts into latent codes and project them to 2D via t-SNE \cite{tsne} for visualization. The plot illustrates that latent codes cluster by clothing and texture, and not by pose, which indicates that pose and clothing are indeed disentangled. 

\subsection{Nearest Neighbor Analysis}
\begin{figure}[]
	\includegraphics[width=0.5\textwidth]{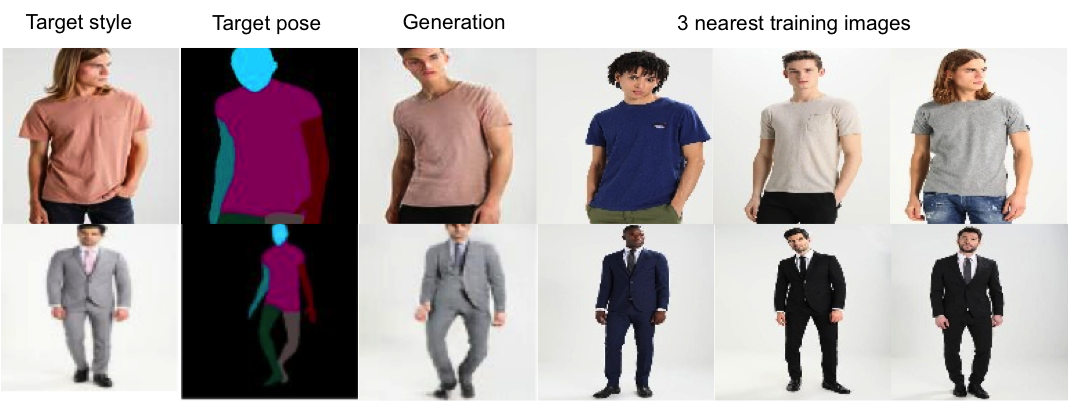}
    \caption{Nearest training images to our generations.}
    \label{fig:nearest}
\end{figure}

A nearest neighbor analysis based on the structural similarity (SSIM) \cite{Wang2004-oz} metric, demonstrates that our learned model does not simply memorize the training data (see Fig.~\ref{fig:nearest}). 

% \subsection{Unpaired Training Data}
% One of the main contributions of our work is the ability to train the network with unpaired data. One can easily add large amounts of real images and , if available, body models, without the need for explicitly pairing or registering these. 
% In practice, availability of real images showing only people in clothing in front of neutral backgrounds may be limited. However, 3D body part models can be rendered via the standard graphics pipeline and are hence attainable in practically unlimited amounts. 
% We study weather additional part models during training improves the image generation results. 
% For this purpose, we train our model with the full body model set and compare it to the same model trained on only half of the training set, with all other parameters staying fixed. 
% Fig.~\ref{fig:adt_data} illustrates that adding additional data improves the overall quality of the results, even for cases where only data for one of the modalities is added.
% \begin{figure}
% 	\includegraphics[width=\textwidth]{pics/partial_data.png}
%     \caption{Each four images form a group. From left to right in a group: real image, part model, synthetic sample from $100\%$ condition, synthetic sample from $50\%$ condition.}
%     \label{fig:adt_data}
% \end{figure}

\begin{figure}
\centering
\includegraphics[width=0.95\columnwidth]{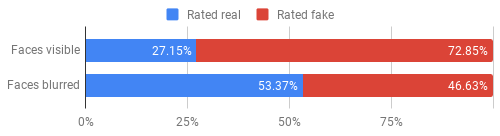}
    \caption{User ratings, without and with blurred faces.\label{fig:user_study}}
    \label{fig:user_ratings}
\end{figure}

\subsection{User Ratings}
Finally, we assess the potentially most important metric to quantify the overall performance of the proposed approach. To better understand if the generated images look realistic we conduct a large-scale perceptual study. 
In this experiment we randomly sample images from the training dataset (\emph{real}) and from synthetic samples generated by our model (\emph{synth}). To isolate the influence of the facial region (which we do not treat specifically), we separate the participants in two groups where the first group judges images with full facial information and the second group judges images with blurred faces. 
% \setlength{\tabcolsep}{4pt}
% \begin{table}
% \caption{Results for Original Images (left) and Face-blurred Images (right)}
% \label{tab:user_study}
% \begin{center}
% \begin{tabular}{ | l | c | c |}
% \hline
%  &  Rated Real &  Rated Gen. \\ \hline
% Generated &          27.15\% & 72.85\% \\ \hline
% Real      &          81.99\% & 18.01\% \\
% \hline
% \end{tabular}
% \quad
% \begin{tabular}{ | l | c | c |}
% \hline
%  &  Rated Real &  Rated Gen. \\ \hline
% Generated &          53.37\% & 46.63\% \\ \hline
% Real      &          68.18\% & 31.82\% \\
% \hline
% \end{tabular}
% \end{center}
% \end{table}
% \setlength{\tabcolsep}{1.4pt}

In total we asked $n=478$ participants to judge $50$ images ($25$ \emph{real} / $25$ \emph{synth}).
In a forced-choice setting, participants had to decide whether an image is real or synthetic. The participants did not know the true distribution of synthetic and real images and were not given any other instructions. Participants were recruited via mailing-lists. 
Fig~\ref{fig:user_ratings} summarizes the results. With faces visible, synthetic images were rated as real with a fool-rate of $\sim 27\%$ which is significantly higher than previous work~\cite{Lassner:GeneratingPeople:2017}. 
When removing the influence of the facial region via face blurring, this result improves to a fool-rate of $\sim 53\%$. This indicates that our model indeed synthesizes realistic samples and improves the perceived image quality over prior work.

\begin{figure}[]
\begin{center}
	\includegraphics[width=0.5\textwidth]{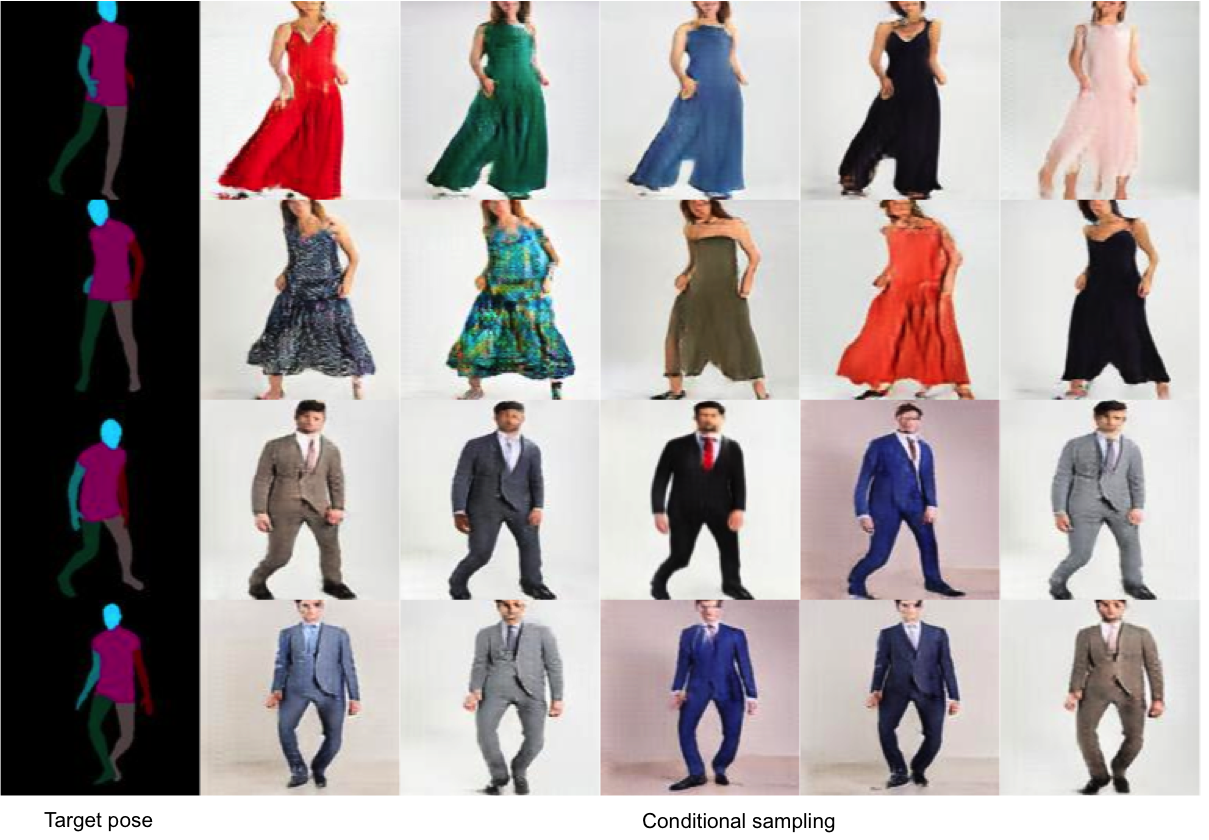}
	\end{center}
    \caption{Samples drawn from the latent space $z$. Each row is conditioned one a particular pose and clothing style.}
    \label{fig:sampl}
\end{figure}

\subsection{Qualitative Results}
\label{sec:exp:ve}
\begin{figure}[]
	\includegraphics[width=0.5\textwidth]{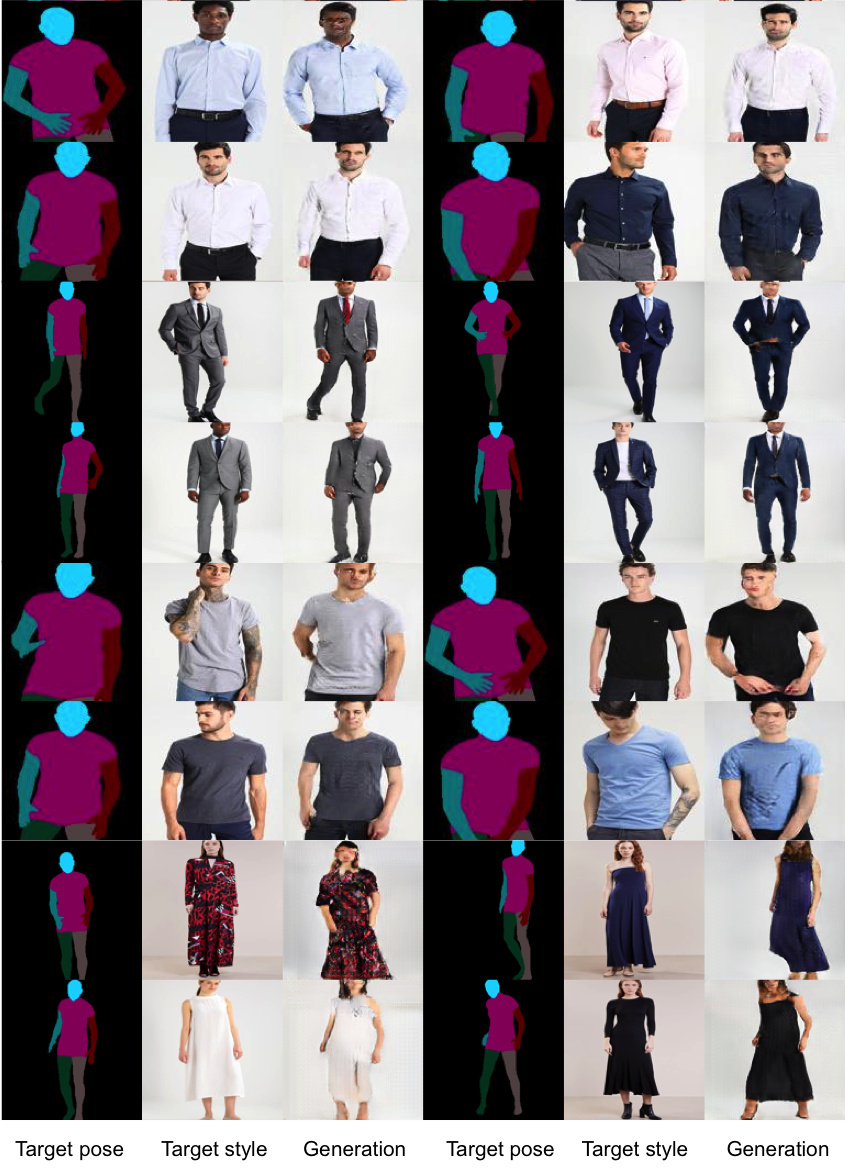}
    \caption{More examples on instance control: provided an image of a person to generate a new image of the person wearing this specific clothing in a particular pose}
    \label{fig:control_more_sample}
\end{figure}

\begin{figure}[]
	\includegraphics[width=0.5\textwidth]{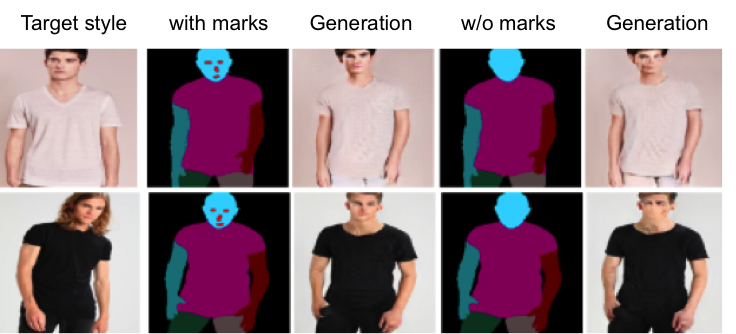}
    \caption{Improving synthetic faces via landmarks.}
    \label{fig:face}
\end{figure}

As discussed in Section~\ref{sec:upg-GAN} and experimentally verified above, the proposed architecture can generate synthetic images, conditioned on either an example image of the target style or a class label. 
%sampling
\figref{fig:sampl} illustrates that the architecture generates images of sufficient quality and is capable of producing samples with significant intra-class variation. 
%instance control
Fig.~\ref{fig:control_more_sample} illustrates both concepts via a number of example sequences where each row shows images generated conditioned on a target pose and a specific clothing style. The samples contain significant diversity in both color and texture, while adhering to the pose guidance given by the input. In terms of image quality, we can see that the synthetic samples accurately capture the body pose information and generally appear realistic. The main challenge stems from the facial regions where neither the part model nor the class label provides any guidance and simultaneously the training data contains a lot of variation. One possibility to alleviate this issue could be to employ facial landmark annotations as shown in Fig.~\ref{fig:face}.

% \begin{figure*}[t!]
% \begin{center}
% 	\includegraphics[width=\textwidth]{pics/sampling.png}
% 	\end{center}
%     \caption{Random samples drawn from the latent space $z$. Each row is conditioned one a particular pose and clothing style.}
%     \label{fig:sampl}
% \end{figure*}

\begin{figure*}[t!]
\begin{center}
	\includegraphics[width=\textwidth]{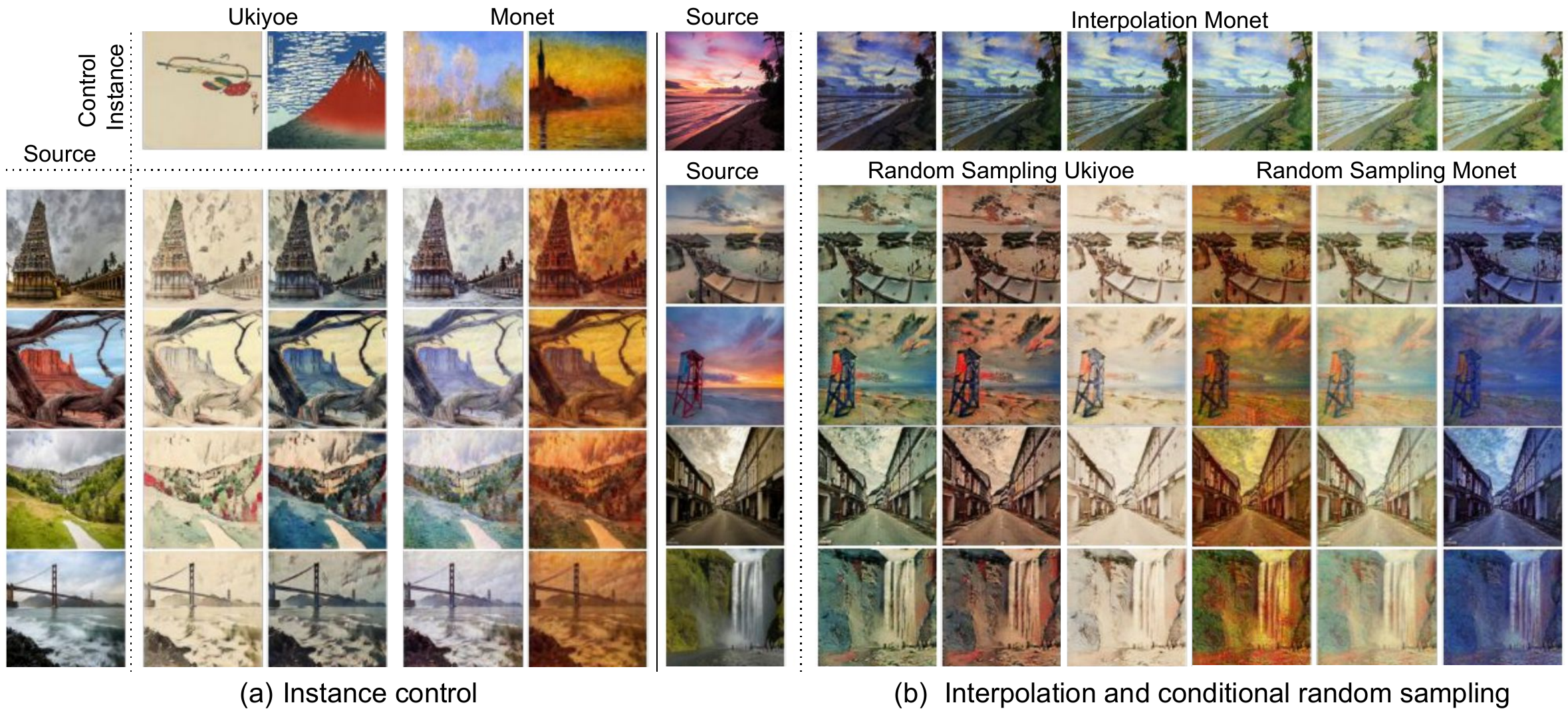}
	\end{center}
    \caption{Qualitative results of photo2painting.}
    \label{fig:painting}
\end{figure*}

\subsection{Limitation and Failure Cases}
\begin{figure}[]
\begin{center}
	\includegraphics[width=0.5\textwidth]{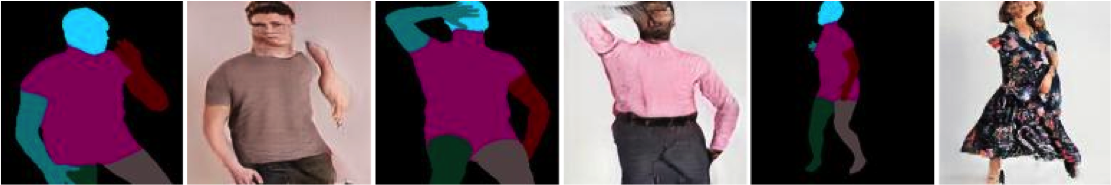}
	\end{center}
    \caption{Failure cases due to uncommon poses or viewing angles.}
    \label{fig:failure}
\end{figure}

Fig.~\ref{fig:failure} depicts several challenging examples and failure cases. The main reason for unnatural synthetic images are uncommon or entirely unseen poses or viewing angles. Due to the unpaired nature of the proposed training regime, this kind of problem may be alleviated via the addition of more diverse training data.

% \begin{figure*}[t!]
% \begin{center}
% 	\includegraphics[width=\textwidth]{pics/failure_cases.png}
% 	\end{center}
%     \caption{Failure cases due to uncommon poses or viewing angles.}
%     \label{fig:failure}
% \end{figure*}

\subsection{Multi-model Conditional Image Translation}
The focus of our work is generative modelling of human images. However, the architecture is general and can also be utilized in other application domains that need to synthesizing images, controlled by guidance that is specified in a different domain. Recent work has tackled the related problem of multi-modal image translation \cite{huang2018multimodal}. We demonstrate in Fig.~\ref{fig:painting} a photo-to-painting example.  We use two styles of paintings and real photos from the dataset proposed in \cite{cyclegan}. Where \cite{huang2018multimodal} train a network per style, we train only one single network for style translation. Furthermore, we show that we can control the style \emph{and} texture of our generated paintings by maintaining content and texture from the input while translating the style from the reference (\ref{fig:painting}, a). Furthermore, we demonstrate the capability to sample from the latent space, varying the texture of the generated image within the desired style (Fig.~\ref{fig:painting}, b). Neither of these two functionalities can be achieved with direct-image-to-image approaches such as CycleGAN.

% !TEX root = ../egpaper_for_review.tex

\section{Conclusion}
In this paper we have introduced the task of purely pose-guided generative modelling of humans in clothing. This task is challenging because it goes beyond the traditional setting of image-to-image translation. Given only a sketch of the human pose as input, we seek to generate realistic looking images that accurately depict the desired pose, and either a specific outfit (red dress), or to generate a number of images that depict the pose and variations of a class of outfit (different dresses). We have contributed a novel architecture that can generate high-quality images in an unpaired setting, while providing either direct instance or class-level control over the depicted style of clothing. We have experimentally shown that the proposed architecture is capable of creating realistic images and an ablative study showed the contributions of the different components of the architecture. A large-scale user-study shows that the generated images are seen as convincing, especially if the faces are blurred. Finally, we have demonstrated that the architecture can also be leveraged for other multi-modal image translation tasks.

Interesting directions for future work include, handling of uncommon poses and generation of images under novel viewpoints. Furthermore, it would be interesting to extend the task to the temporal domain which would also require modelling of the dynamics of the human body and its interactions with the non-rigid deformation of clothing. 

{\small
\bibliographystyle{ieee}
\bibliography{egbib}
}

\end{document}